\documentclass[sigconf,review = False]{acmart}

\usepackage{algorithm}
\usepackage{algorithmic}
\usepackage{microtype}
\usepackage{graphicx}
\usepackage{subfigure}
\usepackage{booktabs} 
\usepackage{amsmath}
\usepackage{bm}
\usepackage{tabularx}
\usepackage{enumitem}
\usepackage{float}

\AtBeginDocument{%
  \providecommand\BibTeX{{%
    \normalfont B\kern-0.5em{\scshape i\kern-0.25em b}\kern-0.8em\TeX}}}

\setcopyright{acmcopyright}
\copyrightyear{2018}
\acmYear{2018}
\acmDOI{10.1145/1122445.1122456}

\acmConference[KDD '21]{Woodstock '18: ACM Symposium on Neural
  Gaze Detection}{August 14--28, 2021}{Virtual Event, Singapore}
\acmBooktitle{Woodstock '18: ACM Symposium on Neural Gaze Detection,
  June 03--05, 2018, Woodstock, NY}
\acmPrice{15.00}
\acmISBN{978-1-4503-XXXX-X/18/06}
\renewcommand\footnotetextcopyrightpermission[1]{} 

\begin{document}
	\newcolumntype{L}[1]{>{\raggedright\arraybackslash}p{#1}}
	\newcolumntype{C}[1]{>{\centering\arraybackslash}p{#1}}
	\newcolumntype{R}[1]{>{\raggedleft\arraybackslash}p{#1}} 
\title{Mutual Information Preserving Back-propagation: \\
Learn to Invert for Faithful Attribution}




 \author{Huiqi Deng, Na Zou, Weifu Chen, Guocan Feng, Mengnan Du, Xia Hu}

\affiliation{
  \institution{Sun Yat-Sen University}
 \city{Guangzhou}
\country{China}
}

\affiliation{
  \institution{Texas A\&M University}
 \city{Guangzhou}
\country{USA}}





\begin{abstract}
Back-propagation based visualizations have been proposed to interpret deep neural networks (DNNs),  some of which produce interpretations with good visual quality. However, there exist doubts about whether these intuitive visualizations are related to the network decisions. Recent studies have confirmed this suspicion by verifying that almost all these modified back-propagation visualizations are not faithful to the model's decision-making process. Besides, these visualizations produce vague “relative importance scores", among which low values can't guarantee to be independent of the final prediction. Hence, it's highly desirable to develop a novel back-propagation framework that guarantees theoretical faithfulness and produces a quantitative attribution score with a clear understanding. To achieve the goal, we resort to mutual information theory to generate the interpretations, studying how much information of output is encoded in each input neuron. The basic idea is to learn a source signal by back-propagation such that the mutual information between input and output should be as much as possible preserved in the mutual information between input and the source signal. In addition, we propose a Mutual Information Preserving Inverse Network, termed MIP-IN, in which the parameters of each layer are recursively trained to learn how to invert. During the inversion, forward relu operation is adopted to adapt the general interpretations to the specific input. We then empirically demonstrate that the inverted source signal satisfies completeness and minimality property, which are crucial for a faithful interpretation. Furthermore, the empirical study validates the effectiveness of interpretations generated by MIP-IN.
\end{abstract}



\keywords{Model interpretation, Back-propagation techniques, Faithfulness, Mutual information preserving}

\maketitle

\section{Introduction}
\label{Introduction}
Attribution~\cite{samek2020toward,fong2017interpretable,lundberg2017unified} is an efficient computational tool to interpret DNNs, inferring the importance score of each input feature to the final decision for a given sample. Back-propagation based visualization~\cite{zeiler2014visualizing,springenberg2014striving,bach2015pixel,montavon2017explaining,sundararajan2017axiomatic,shrikumar2017learning,kindermans2017learning} is a classic approach in attribution interpretations, which propagates an importance score from the target output neuron through each layer to input neurons in one pass. Different propagation rules are adopted in existing work.
For example, deconvnet~\cite{zeiler2014visualizing} maps the output neuron back to the input pixel space by resorting to backward relu and transposed filters in a layer-wise manner. Deep Taylor decomposition ~\cite{montavon2017explaining} hierarchically decomposes each upper layer neuron to lower layer neurons according to the upper layer neuron’s first-order Taylor expansion at the nearest root point.
Back-propagation based methods are usually efficient, parameter-free, some of which produce interpretations with good visual quality and achieve good performance on image degradation and object localization task. Besides, back-propagation methods enable fine-grained visualizations and layer-wise explanations, which may provide more comprehensive
understandings to the decision-making process of the neural network than the perturbation-based attribution methods~\cite{fong2017interpretable, fong2019understanding}.

Despite the human-interpretable visualizations,
several works have raised doubts about the faithfulness of these back-propagation visualizations recently ~\cite{adebayo2018sanity,nie2018theoretical,sixt2020explanations}. \emph{Do the pretty visualizations really reliably reflect why the network makes such a decision, e.g., classify an image as a cat?}
There always exists a concern that the theoretical rationales of most back-propagation methods are confusing and hard to explore~\cite{nie2018theoretical,kindermans2017learning}. For example, the original purpose of the transposed filter and backward relu in Deconvnet is to preserve the neuron size and ensure the non-negative neurons in intermediate layers~\cite{zeiler2014visualizing}, which have no direct connections to interpreting the final prediction. In another example, the Guided Back-propagation (GBP)~\cite{springenberg2014striving} significantly improves the visualization by merely combining the forward relu of saliency map and the backward relu of Deconvnet. The interpretations based on those perplexing behaviors can not be fully trusted.

Recent studies have further confirmed the above-mentioned suspicions and verified that \emph{most of the modified back-propagation interpretations are not faithful to the model's decision-making process}. A theoretical analysis has revealed that Deconvnet and GBP are essentially doing (partial) image recovery which is unrelated to the network decisions~\cite{nie2018theoretical}. Besides,
Some empirical evidence have also been proposed. It's considered a method insensitive to the class labels and parameters can not explain the network’s prediction faithfully~\cite{nie2018theoretical, adebayo2018sanity}.
However, the previous observations
~\cite{samek2016evaluating,selvaraju2017grad,mahendran2016salient} have shown that the visualizations of Deconvnet and GBP keep almost the same given different class labels. The investigations and conclusions have been extended to most of back-propagation based attribution methods, including Deep Taylor decomposition~\cite{montavon2017explaining}, Layer-wise Relevance Propagation (LRP-$\alpha\beta$) ~\cite{bach2015pixel}, Excitation BP~\cite{zhang2018top}, PatternAttribution~\cite{kindermans2017learning}, Deconvnet, and GBP. The investigations demonstrate that the attributions of these methods are independent of the parameters in the layers of classification module~\cite{sixt2020explanations} and hence fail the sanity check~\cite{adebayo2018sanity}. In addition, existing back-propagation attribution methods \emph{generate a vague ``relative importance score", in which the numerical values have no practical implications}. Specifically, the low relevance values do not guarantee the network will ignore the corresponding part during the decision-making process~\cite{Schulz2020Restricting}.

Thus, it is highly desirable to develop a novel back-propagation framework that guarantees  theoretical faithfulness and produces a quantitative attribution score with a clear understanding.
To achieve these two goals, we resort to the mutual information theory, which is capable of generating sound interpretations and estimating the amount of information. We aim to study how much information of prediction is encoded in each input neuron around the neighborhood of a given sample. It's intractable to directly deal with the mutual information for each sample, which is notoriously hard to estimate.
Motivated by mask-learning methods ~\cite{fong2017interpretable} which learns a masked input maximally preserving the discriminative information, our basic idea is to learn a source signal via back-propagation. The source signal should satisfy that the mutual information between input and output should be preserved as much as possible in the mutual information between input and the source signal. We further formulate this objective as minimizing the conditional entropy of the input given the source signal, which could be approximately achieved via the reconstruction-type objective function (Fano's inequality). Then to solve the optimization problem, we propose Mutual Information Preserving Inverse Network, termed as MIP-IN, a layer-wise inverse network. MIP-IN recursively retrain the parameters of each layer to learn how to invert and then adopt forward relu to adapt the global framework to each specific input.

We conduct experiments to validate the effectiveness of the proposed MIP-IN framework. Firstly, we empirically validate that the source signal produced by MIP-IN satisfies two good properties: completeness and minimality. Secondly, we show that MIP-IN generates human-interpretable visualizations for models without local convolution, while other back-propagation based methods fail to do so. Thirdly, we demonstrate that MIP-IN produces visualizations of good quality, which well locates the object of interest. Finally, we investigate the sensitivity of interpretations w.r.t the class labels. We observe that MIP-IN is very sensitive to the change of class labels, while most of modified back-propagation attribution methods are not.
In summary, this paper has three main contributions:
\begin{itemize}[leftmargin=*]
    \item We formulate the back-propagation attribution problem as learning a source signal in the input space maximally preserving the mutual information between input and output.
    \item We propose a novel mutual information preserving inverse network, termed MIP-IN, to learn how to invert for attribution.
    \item Experimental results on two benchmark datasets validate that the proposed MIP-IN framework could generate attribution heatmaps that have high quality and are faithful to the decision-making process of underlying DNN.
\end{itemize}

\section{Related work}
In this section, we provide an overview of existing attribution methods that could infer the importance of each input feature to the final prediction for a given input sample.

\begin{figure}[t]\centering
\includegraphics[scale=0.42]{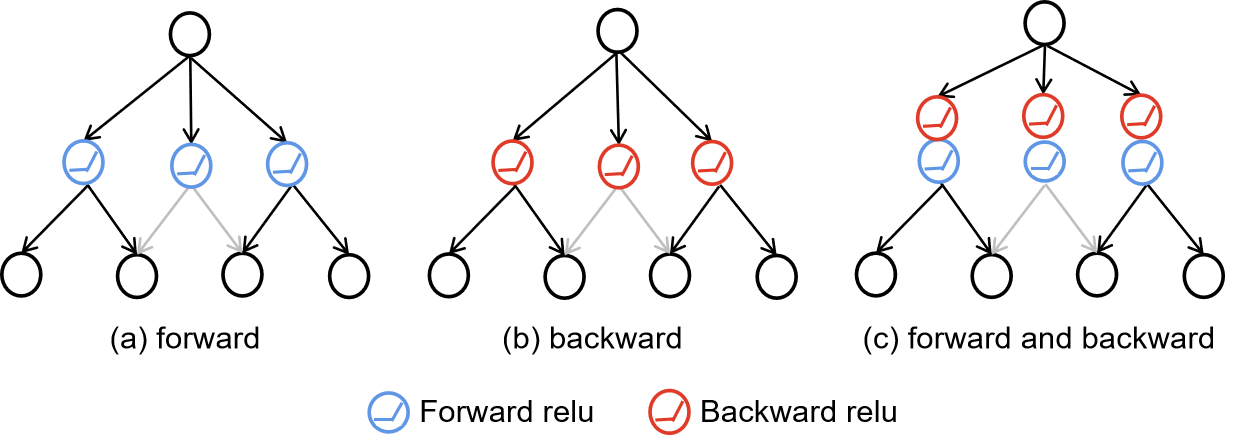}
\caption{Summary of existing back-propagation structures. Grads, Smooth grads, Integrated Grads, LRP, DTD, and PatternAttribution adopt the forward structure. Deconvnet adopts the backward structure, while GBP adopts the forward and backward structure.  }
\vspace{-4pt}
\label{Related work}
\end{figure}

\subsection{Back-Propagation Based Attribution}
Back-propagation based attribution is an efficient attribution approach that propagates an importance score from an output neuron through each layer to the input neurons in one pass. Many different back-propagation methods are proposed based on different back-propagation rules.

Gradient~\cite{baehrens2010explain} describes the sensitivity of the change of the output neuron w.r.t each input neuron, which masks out the negative neurons of bottom data via the forward Relu.  Deconvnet\cite{zeiler2014visualizing} aims to map the output neuron back to the input pixel space. To keep the neuron's size and non-negative property, they resort to the transposed filters and backward relu, which masks out the negative neurons of the top gradients. The Guided Back-propagation (GBP) ~\cite{springenberg2014striving} combines Gradients and Deconvnet, considering both forward relu and backward relu. As a result, GBP could significantly improve the visual quality of visualizations.

In addition, other back-propagation attributions follow the idea of directly or hierarchically decomposing the value of the target neuron to the input neurons.
Layerwise Relevance Propagation (LRP) ~\cite{bach2015pixel} decomposes the relevance score to the neurons of the lower layer according to the corresponding proportion in the linear combination. DeepLift ~\cite{shrikumar2017learning} adopts a similar linear rule, while it assigns the difference between the output and a reference output in terms of the  difference between the input and a pre-set reference input, instead of merely using the output value. Integrated Gradients ~\cite{sundararajan2017axiomatic} also decomposes the output difference by integrating the gradients along a straight path interpolating between input sample and a reference point, which corresponds to \emph{Aumann Shapley} payoff assignment in cooperative game theory. Smooth Gradients ~\cite{smilkov2017smoothgrad} is similar to Integrated Gradients, which averages the gradients in a local neighborhood sampled by adding Gaussian noise.
Deep Taylor decomposition (DTD)~\cite{montavon2017explaining} hierarchically decomposes each upper layer neuron to the lower layer neurons according to the upper layer neuron’s first-order Taylor expansion at the nearest root point. PatternAttribution~\cite{kindermans2017learning} extends the DTD by training a function from data to learn the root points. Their different back-propagation structures are summarized in Figure \ref{Related work}.

Nevertheless, recent studies ~\cite{nie2018theoretical, adebayo2018sanity,samek2016evaluating,selvaraju2017grad,mahendran2016salient,Schulz2020Restricting} have indicated that most modified back-propagation based interpretations are not faithful to the model's decision-making process. This motivates us to develop a novel back-propagation framework that guarantees theoretical faithfulness and produces a quantitative attribution score with a clear understanding.

\subsection{Perturbation-Based Attribution}
Apart from the back-propagation based attribution approach, many perturbation-based attribution methods have been proposed. They generate a meaningful explanation by informative subset learning for CNNs~\cite{fong2017interpretable,dabkowski2017real,wagner2019interpretable,Schulz2020Restricting} and GNNs~\cite{ying2019gnnexplainer}. The basic idea behind is to maximize the mutual information between the predictions $Y$ of a neural network and the distribution of possible subset structure $X_s$. For a pre-trained neural network $\Phi$, $X$ and $Y = \Phi(X)$ denote its input variable and corresponding final prediction variable. Then the informative subset learning problem can be formulated as:
\begin{equation}\nonumber
    max_{X_s} MI(X_s,Y) \quad st \ |X_s|< M,
\end{equation}
so as to find the most informative subset $X_s$ for the prediction $Y$.

To make $X_s$ most compact, a constraint on the size of $X_s$ or sparse constraints are usually imposed. Noted that $MI(X_s, Y) = H(Y) - H(Y|X = X_s)$ and the entropy term $H(Y)$ is constant because $\Phi$ is fixed for a pre-trained network. As a result, maximizing the mutual information $MI(X_s,Y)$ is equivalent to minimizing the conditional entropy $H(Y|X = X_s)$, that is:
\begin{equation}
    min_{X_s} H(Y|X = X_s) \quad st \ |X_s|< M.
\end{equation}
The distribution of possible subset structure $X_s$ for a given input sample $x$ is often constructed by introducing a Gaussian noise mask to the input. Specially, the input variable $X_s = m x + (1-m)\eta$, in which the Gaussian noise $\eta \sim \mathcal{N}(0,\sigma^2)$ and the mask vector $m$ needs to be optimized to learn the most informative subset. Besides, when the end-users come to the question "why does the trained model predict the instance as a certain class label", the above conditional entropy $H(Y|X_s)$ is usually modified by the cross entropy objective between the label class and the model prediction.

Informative subset learning methods generate meaningful explanations, but suffer from some drawbacks: i) they incur the risk of triggering artifacts of the black-box model. ii) the produced explanations are forced to be smooth, which prevents fine-grained evidence from being visualized.

\section{Learn to invert with mutual information preserving}
In this section, we first introduce the instance-level attribution problem that we aim to tackle, and then discuss the mutual information preserving principle that a back-propagation based attribution should follow in order to generate faithful interpretations.

\subsection{Problem Statement}
To interpret a pre-trained network $\Phi$, the attribution method targets at inferring the contribution score of each input feature to the final decision $y$ for a given sample $x = [x_1, \dots, x_n]$~\cite{du2019techniques}. Back-propagation based attribution is an efficient and parameter-free attribution approach, which usually enables fine-grained explanations.
Different from existing back-propagation based methods, we develop a novel Back-propagation explanation method from the mutual information perspective. Specifically, we try to solve the attribution problem by answering the following two research questions: i) In a global view, for input variable $X$ and output variable $Y = \Phi(X)$, how much information of $Y$ is encoded in each input feature? ii) In a local view, for a specific input $x$ to be explained, how much information of $Y$ in a local neighbor of $x$ is encoded in each input feature?

\subsection{Problem Formulation}
\label{Problem formulation}
The informative subset learning approach~\cite{fong2017interpretable} learns a masked input which masks out the irrelevant region to maximally preserves the information related to the final prediction. Motivated by the approach, we answer the first question by inverting the output variable back to the input (pixel) space with the mutual information preserving principle. Specifically, we aim to learn a set of source signal $S = \{s\}$ and distractor signal $D = \{d\}$ of input signal $X = \{x\}$ in the input space, which satisfy that each input signal $x$ in $X$ is composed of its corresponding source signal $s$ and distractor $d$, i.e., $x = s + d$.  Here, $s$ is the desired inverted signal.
We hope the inverted source signal $s$ contains almost all the signal of interest correlating with final prediction $y$, while the distractor signal $d$ represents the remaining part in which no information of $y$ can be captured. More formally, the mutual information between input $X$ and output prediction $Y$ should be preserved as much as possible in the mutual information between $X$ and $S$, not in $D$. Hence, the inverted source signal $S$ could be optimized by:
\begin{equation}
    min_{S} \ |MI(X,S) - MI(X,Y)|.
\end{equation}
It can be observed that the inverted source signal $S$ can be expressed as a function of the prediction $Y$, i.e., $S = g(Y)$, when no extra variable except for $y$ is introduced during the back-propagation process. According to the data processing inequality of mutual information, we have $MI(X,S) \leq MI(X,Y)$, which means the source signal $S$ does not introduce the information out of the prediction $Y$. The term $MI(X,Y)$ is constant for a pre-trained network,
hence maximally preserving the mutual information between input $X$ and output $Y$ is equivalent to maximizing the mutual information between $X$ and the source signal $S = g(Y)$, i.e.,
\begin{equation}
    max_{g} MI(X,g(Y)) =  H(X) - H(X|g(Y)).
    \label{maximizeMIXS}
\end{equation}
Because the entropy $H(X)$ of input is constant, optimizing equation (\ref{maximizeMIXS}) is equivalent to minimizing the conditional entropy $H(X|g(Y))$, which can be expressed as:
\begin{equation}\nonumber
   min_g  H(X|g(Y)) = - E_{g(Y)} log(P(X|g(Y))).
\end{equation}
It's intractable to compute $H(X|g(Y))$  since we need to integrate $P(X|g(Y))$ over the distribution of $g(Y)$. Hence, instead of directly optimizing $H(X|g(Y))$, we use a reconstruction-type objective to approximately optimize such conditional entropy, whose feasibility is guaranteed by the following Theorem 1.

\vspace{3pt}
\noindent \textbf{Theorem 1.}
\emph{Let the random variables $X$ and $Y$ represent input and output messages with a joint probability $P(x,y)$. Let $\epsilon$ represent an occurrence of reconstruction error event, i.e., that $X \neq \tilde X$ with $\tilde X = g(Y)$ being an approximate version of $X$. We have Fano's inequality,
\begin{equation}\nonumber
    H(X|\tilde X) \leq H(\epsilon) + P(\epsilon) \ log(|\mathcal{X}| -1),
\end{equation}
where $\mathcal{X}$ is the alphabet of $X$, and $|\mathcal{X}|$ is the cardinality of  $\mathcal{X}$. $P(\epsilon) = P(X \neq \tilde X)$ is the probability of existing reconstruction  error and $H(\epsilon)$ is the corresponding binary entropy.}

The proof of Theorem 1 is widely accessible in standard references such as~\cite{cover1999elements}. The Fano's inequality states the relationship between the conditional entropy $H(X|\tilde X)$ and the probability of reconstruction error $P(\epsilon) = P(X \neq \tilde X)$. The inequality demonstrates that minimizing the conditional entropy $H(X|\tilde X)$ can be approximately achieved by minimizing its upper bound, i.e., minimizing the probability $P(\epsilon)$. Hence it's feasible to minimize the reconstruction error between $X$ and $g(Y)$ to guarantee the minimization of the amount of information of $g(Y)$ encoded in $X$. Assume $X$ and $g(Y)$ are normally distributed, the probability of reconstruction error should be minimal by minimizing the mean square error $MSE(X, g(Y))$ between $X$ and $g(Y)$, that is,
\begin{equation}\label{Objective}
    min_{g} MSE(X,g(Y)) = ||X - g(Y)||^2.
\end{equation}
Then the mutual information $MI(X,S)$ between input $X$ and its source signal $S$ can characterize the importance of each input feature to the final prediction $Y$ globally.

It becomes a little different when the explanation comes to a local view for a specific input $x$. The above formulation emphasizes the features of interest globally, which averages over the whole data distribution. While for a specific input, we are more interested in the model behavior around its local neighborhood. Hence, we need to adjust such global explanations to the specific input. We conduct the adaptation by incorporating some prior information about the neighborhood of $x$.
We will elaborate on the detailed methodology of the local adaptation in section \ref{specific}.

\section{MIP-IN: Mutual Information Preserving Inverse Network}
\begin{figure}[t]\centering
\includegraphics[scale=0.38]{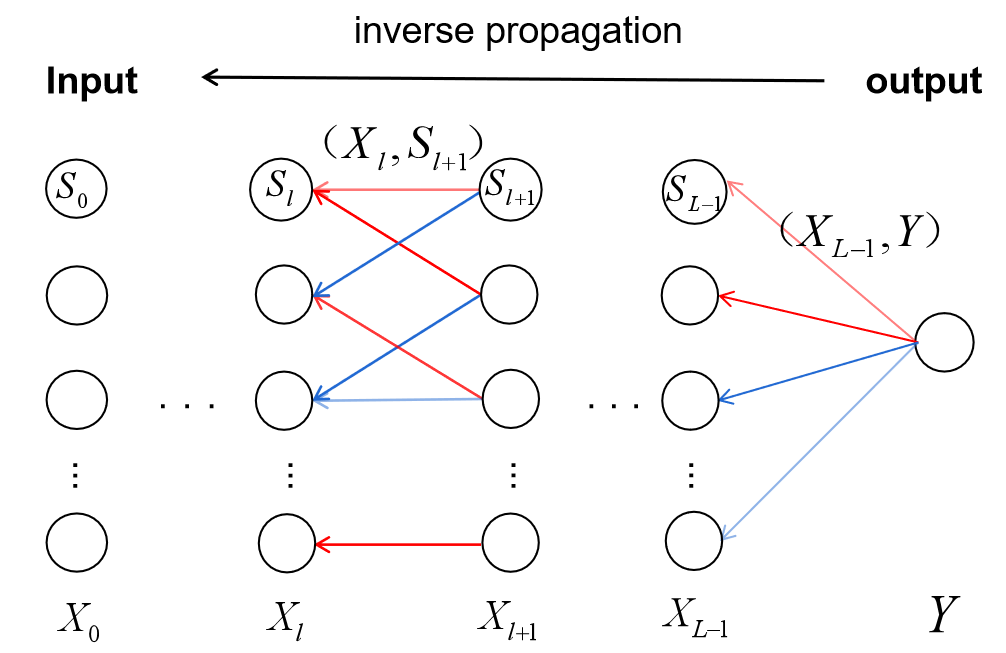}
\caption{Computational flow of MIP-IN framework. The inverted source signal $S_0$ is recursively propagated from the output layer to the input layer, and $S_l$ represents the intermediate inverted signal at $l$-th layer. Red and blue arrows represent positive and negative parameters, respectively. The symbol $(X_l, S_{l+1})$ above the parameter means the parameter of MIP-IN at $l$-th layer is learned by using the inverted source signal $S_{l+1}$ to reconstruct the original signal $X_l$ at $l$-th layer. Similarly for the symbol $(X_{L-1}, Y)$. }
\label{Computational flow}
\end{figure}

In this section, we propose Mutual Information Preserving Inverse Network (MIP-IN), which is a layer-wise inverse network (Figure \ref{Computational flow}) based on the mutual information preserving principle introduced in the previous section.

In MIP-IN, the mutual information between input and output prediction is maximally preserved in the inverted source signal during back-propagation. Assume the DNN has $L$ layers, we construct the inverse function $g$ as a multiple composite function, i.e.,
\begin{equation}\nonumber
\begin{aligned}
    & g = g_1(g_2\dots g_{L-1}(g_{L})), \\
    & \text{Then} \quad S = g(Y),
\end{aligned}
\end{equation}
where $Y$ denotes the output neuron, and $g_l$ denotes the inverse function at $l$-th layer. And the source signal $S$ is obtained via layer-wise back-propagation. According to the objective of equation (\ref{Objective}) in section \ref{Problem formulation}, we learn the inverse function by minimizing the L2 norm of the difference between input $X$ and the source signal $S$,
\begin{equation}\nonumber
    min_{\{g_1, \dots, g_L\}} ||X - g_1(g_2\dots g_{L-1}(g_{L}(Y)))||^2.
\end{equation}
To guarantee the faithfulness to the model, we further restrict that the mutual information between every two adjacent layers is also maximally preserved during the inversion. Hence we recursively learn the inverse function from up layers to bottom layers. We initialize $S_L = Y$. Then, $g_L(Y)$ aims to reconstruct the neuron entries at $L-1$ layer, which is expressed as:
\begin{equation}\nonumber
\begin{aligned}
     & min_{\{g_{L}\}} ||X_{L-1} - g_L(Y)||^2   \\
     & \text{Then} \quad S_{L-1}  = g_L(Y),
\end{aligned}
\end{equation}
where $X_l$ denotes the intermediate feature map at the $l$-th layer, and $S_l$ represents the corresponding source signal of $X_l$ at $l$-th layer. Similarly, at $l$-th layer with $g_L, g_{L-1},\dots, g_{l+2}$ obtained,  $g_{l+1}(S_{l+1})$ aims to reconstruct the neuron entries at $l$-th layer, i.e.,
\begin{equation}
\label{MIP-IN framework}
\begin{aligned}
    & min_{\{g_{l+1}\}} ||X_{l} - g_{l+1}(S_{l+1})||^2 \\
    & \text{Then} \quad S_{l} = g_{l+1}(S_{l+1}).
    \end{aligned}
\end{equation}
The computational flow of MIP-IN is shown in Figure \ref{Computational flow}.

In the following, we further introduce the proposed MIP-IN framework in details. Firstly, we design the type of inverse function $g$ for different types of layers. The parameter of $g_l$ at each layer can be optimized by minimizing the reconstruction error. Then, we elaborate how to leverage the activation switch information to adapt to a specific input sample $x$. Finally, we introduce how to generate the interpretations based on MIP-IN.
The overall pipeline is summarized in Algorithm \ref{alg:LayerwiseInverse}, where the back-propagation operation follows the forward structure in Figure \ref{Related work}.

\subsection{Design of The Inverse Function}
In this subsection, we design the function $g$ for (i) dense, (ii) convolution, and (iii) max-pooling layer, respectively. We focus on the block at $l$-th layer, which uses the intermediate inverted source signal $S_{l+1}$ to reconstruct $X_l$, as shown in Equation (\ref{MIP-IN framework}). For simplicity, we write the neurons at $l$-th layer $X_{l}$ and the inverted signal at $(l+1)$-layer $S_{l+1}$ as $X$ and $S$.

\vspace{2pt}
\noindent\textbf{Dense Layer:}
For $l$-th dense layers, we define the corresponding back-propagation inverse function $g^D(Y)$ as a linear function, which can be expressed as $g^D(Y) = WY + b$. According to the recursive reconstruction objective, we need to minimize the reconstruction error between $X$ and $g^D(S)$, i.e., $||X - (WS + b)||^2$. Then $W$ and $b$ have an unique closed solution:
\begin{equation}\label{Dense layer}
\begin{aligned}
 W &= (\bar X\bar S^T)(\bar S \bar S^T)^{-1}\\
    b &= E(X) - AE(S),
\end{aligned}
\end{equation}
where $\bar X = X - E(X)$ and $\bar S = S - E(S)$ represent the centralized input and inverted source signal respectively. To avoid over-fitting, we adopt a L2 norm to regularize the parameters $W$. Then the solution becomes $W = (\bar X\bar S^T)(\bar S \bar S^T +\lambda I )^{-1}$, where $\lambda$ controls the trade-off between reconstruction error and L2 norm regularization.

\vspace{2pt}
\noindent\textbf{Convolutional Layers:}
The convnet uses learned filters to convolve the feature maps from the previous layer. It has shown that transposed filter structure can partially recover the signal~\cite{nie2018theoretical}. Besides, such structure accurately describes the correlation relationships, i.e., which neurons at lower layer will be related to which neurons at next upper layer.
Hence we set the transposed filter structure $g^C$  as inverse functions for convolutional layers, where the kernel parameter of $g^C$ can be obtained by minimizing the reconstruction objective function, i.e.,
\begin{equation}\label{Convolution layer}
    min_{g^C} ||X - (g^C * S)||^2.
\end{equation}
Noted that it's different from Deconvnet and GBP, which directly use the transposed version of the same filter in original network.

\vspace{2pt}
\noindent\textbf{Max-pooling Layers:}
We follow the design of Deconvnet on the max-pooling operation, which obtains an approximate inversion by recording the locations of the maximum within each pooling region in a set of switch variables. We use these recorded switches to place the intermediate inverted signal into appropriate locations, preserving the information. The operation $g^M$ is formalized as:
\begin{equation}\label{Max-pooling layer}
\begin{aligned}
     & \text{switch} = record(X_{max}), \\
     & g^M(S,\text{switch})  = S \odot \text{switch}.
\end{aligned}
\end{equation}
Here $\odot$ represents element-wise operation.

\begin{algorithm}[t]
\renewcommand{\algorithmicrequire}{\textbf{Input:}}
\renewcommand{\algorithmicensure}{\textbf{Output:}}
\caption{Mutual Information Preserving Inverse Network}
\begin{algorithmic}[1]
\REQUIRE Input neurons $X_0$, output neurons $Y$, and the intermediate layer neurons $\{X_1,\dots, X_{L-1}\}$ for $N$ samples. The target neuron index of interest $c$.
\STATE \emph{Initialization}: $S_L = Y_c, A = Y_c$;
\FOR {$l = L, L-1, \dots,1$}
    \STATE \emph{feed} \ $X_{l}, S_{l+1}$;
    \IF{\text{Dense layer}}
    \STATE $g_{l+1} = \text{Eq.} (\ref{Dense layer})$;
    \ELSIF{\text{Convolutional layer}}
    \STATE  $g_{l+1} = \text{Eq.} (\ref{Convolution layer})$;
    \ELSIF{Max-pooling layer}
    \STATE  $g_{l+1} = \text{Eq.} (\ref{Max-pooling layer})$;
    \ENDIF
    \STATE $S_l = g_{l+1}(S_{l+1})$;
    \STATE $A_l = exact(g_{l+1})$;
    \STATE $A = A*A_l$;
    \STATE $\text{Indication} = \text{Eq.}(\ref{activation switch})$;
    \STATE \emph{Forward relu}: $S_l = S_l \odot \text{Indication}$;
    \STATE \emph{Forward relu}: $A = A \odot \text{Indication}$;
\ENDFOR
\STATE \textbf{return} the parameters of the inversion function $g_l$ for each layer, the inverted source signals $S_0$, and the attribution vectors $A$ for $N$ samples
\end{algorithmic}
\label{alg:LayerwiseInverse}
\end{algorithm}

\subsection{Forward Relu for Local Adaptation}
\label{specific}
The above framework emphasizes the features of interest globally, which averages over the whole data distribution. It contains no specific information for a given sample. While for a specific input $x$, we are interested in how much information of $Y$ around its local neighborhood is encoded in each input feature. Hence, we need to adapt the global framework to explain a specific input.

The basic idea is to explore and leverage the neighborhood information of $x$. An intuitive way is to draw neighborhood samples according to the prior distribution of $Y$, and train the inversion function on those samples. However, such an approach is clumsy because we need to retrain the network parameters for each sample to be interpreted, which is not scalable in a large DNN.

Instead, we adopt a shortcut by leveraging the activation switch information, i.e., whether the neurons at intermediate layers are activated.
We find that the activation switch, is an effective indicator of neighborhood information. Specifically, we consider the $i$-th neuron $x_l(i)$ at $l$-th layer of the sample $x$, which is obtained by $x_l(i) = relu(W_i x_{l-1} + b_i)$. $W_i$ and $b_i$ denote the corresponding weights and bias respectively. Non-activated neuron $x_l(i) = 0$ implies that the linear combination
$W_i x_{l-1} + b_i <0$. According to continuity assumption, the linear combination would be smaller than $0$ with a high probability for the samples $x'$ whose intermediate feature map $x'_{l-1}$ at $(l-1)$-th layer is near to the one of $x$. Therefore, for these neighboring samples, the corresponding neurons $x_{l}^{'}$ would also be non-activated with a high probability.

Hence, we leverage the activation switch information at each layer to adapt the global framework to the specific input $x$. Specifically, we adopt forward relu, i.e., we force the locations which are non-activated in the original neurons are also non-activated in the inverted source signals as above-mentioned.
\begin{equation}\label{activation switch}
\begin{aligned}
& \text{Indication}(i) =\left\{
\begin{aligned}
1  \quad & \text{if} \ X_l(i) \neq 0 \\
0  \quad & \text{if} \ X_l(i) = 0 \\
\end{aligned}
\right.\\
&S_l = S_l \odot \text{Indication}.
\end{aligned}
\end{equation}
The forward relu operation could well recover the neighbor information of the specific input $x$.

\begin{figure}[t]\centering
\includegraphics[scale=0.55]{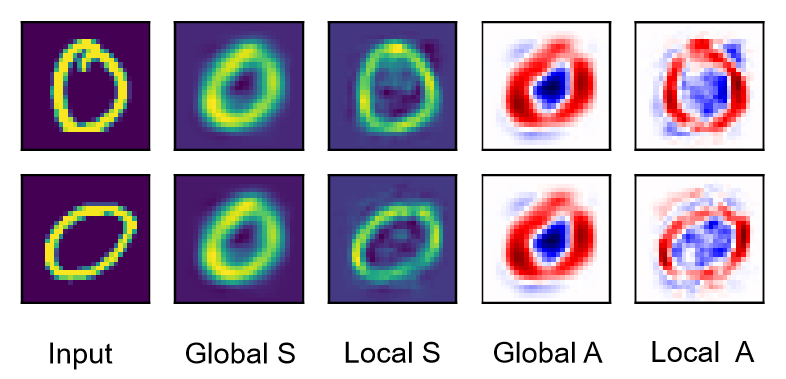}
\caption{Globally and locally produced source signals $S$ and attribution vectors $A$ of samples.}
\vspace{-2pt}
\label{With or without forward relu}
\end{figure}

\subsection{Generating Interpretations}
We generate the interpretations by studying how much information of output neuron $Y$ is encoded in the input neurons $X$ via source signal $S = g(Y)$, which maximally preserves the mutual information between input $X$ and output $Y$.

As the inverse function $g^D$ and $g^C$ for dense layer and convolutional layer are essentially linear functions, we could exact the corresponding linear weight as the attribution $A_l$ at $l$-th layer and ignore the irrelevant bias terms (Algorithm \ref{alg:LayerwiseInverse} line 12). The attribution $A$ is calculated as a superposition of $A_l$ in each layer (Algorithm \ref{alg:LayerwiseInverse} line 13). When adapting to a specific input, we adopt a forward relu operation as in section \ref{specific} (Algorithm \ref{alg:LayerwiseInverse} line 16). \\

The effect of local adaptation is shown in Figure \ref{With or without forward relu}. We can observe that without forward relu, the produced global source signal and attribution vector keep almost the same for different samples in the same class. This is mainly because the global framework obtains an average over the whole data distribution. While after adopting the forward relu function for local adaptation, the framework can well adapt to each specific input. Besides, the attributions of local adaptation turn out to be specifications of the global attributions, emphasizing similar parts of digit $0$.

\begin{table}[t]
\centering
\small
\renewcommand\arraystretch{1.5}
\begin{tabular}{ C{2.0cm}|C{1.5cm}|C{1.6cm}|C{1.6cm}} \hline
\textbf{Models} &  \textbf{MLP-M} & \textbf{CNN-M}  & \textbf{CNN-C} \\  \hline
APC & 10.10\% &  16.5\%  & 21.1\%\\ \hline
Positive APC & 2.6\% &  2.3\%  & 10.6\%  \\ \hline
\end{tabular}
\setlength{\abovecaptionskip}{0pt}%
\setlength{\belowcaptionskip}{4pt}%
\caption{APCs $\epsilon$ and Positive APCs $\epsilon^+$ of the source signal. MLP-M, CNN-M, and CNN-C represent MLP model trained on MNIST, CNN model trained on MNIST, and CNN model trained on Cifar-10 dataset respectively.}
\label{Average Percentage Change}
\end{table}

\begin{figure}[t]\centering
\includegraphics[scale=0.46]{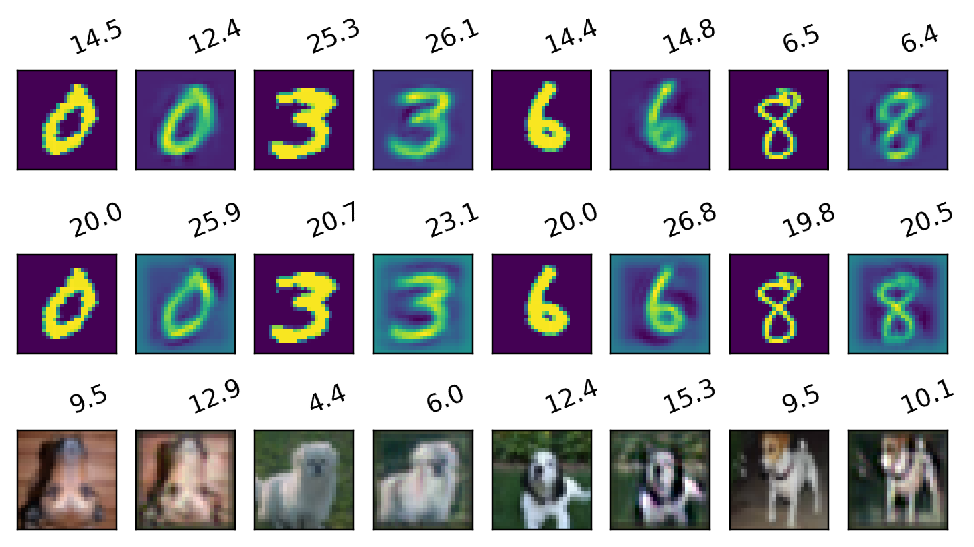}
\caption{Samples and their source signals with logits showing in title. The top, middle, bottom row visualize the results of MLP-M, CNN-M and CNN-C models respectively.}
\vspace{-4pt}
\label{Completeness visualization}
\end{figure}

\section{Evaluation of the Source Signal}
In this section, we introduce how to evaluate the performance of the source signals. It's expected that the source signal $S$ should satisfy two properties: i) \textbf{Completeness:} $S$ preserves almost all the information of output $Y$
and the distractor signal $D$ captures no relevant information of $Y$
, which means $MI(S,Y) \approx MI(X,Y)$; ii) \textbf{Minimality:} no redundant information (irrelevant with $Y$) is obtained in $S$, which means $MI(X,S) \leq MI(X,Y)$.

The produced source signal $S$ naturally well satisfies the minimality because the construction of $S$ adopts a back-propagation manner, in which almost no extraneous information is introduced. Hence, we focus on evaluating the completeness property. $MI(S,Y)$ should be close to $MI(X,Y)$ when the conditional entropy $H(Y|S)$ is close to $H(Y|X)$. The two conditional entropies are equivalent when the output of $S$ is the same as the output of $X$, i.e., $\Phi(S) =\Phi(X)$. Here $\Phi(X), \Phi(S)$ represent the logits of $X, S$ in the network. We adopt the average percentage change (APC) as the metric to measure the difference between $\Phi(S)$ and $\Phi(X)$,
\begin{equation}
\label{APCs}
    \epsilon = \frac{1}{C} \sum_{c = 1}^C \frac{1}{N_c}\sum_{i = 1}^{N_c} \frac{|\Phi(X^i) - \Phi(S^i)|}{|\Phi(X^i)|} \times 100\%,
\end{equation}
where $X^i, S^i$ represent the input and inverted source signal of $i$-th sample. $C$ is the number of classes and $N_c$ is the number of samples in the $c$-th class. We also propose a relaxed metric, termed positive average percentage change (Positive APC), which uses relu function to replace the previous absolute function,
\begin{equation}
\label{positive APCs}
      \epsilon^+ = \frac{1}{C} \sum_{c = 1}^C \frac{1}{N_c}\sum_{i = 1}^{N_c} \frac{relu(\Phi(X^i) - \Phi(S^i))}{|\Phi(X^i)|} \times 100\%.
\end{equation}
Because it can be considered that $S^i$ contains the sufficient information for prediction when $\Phi(X^i) < \Phi(S^i)$.
The completeness is satisfied when $\epsilon,\epsilon^+$ are very small.

\begin{figure}[t]\centering
\includegraphics[scale = 0.35]{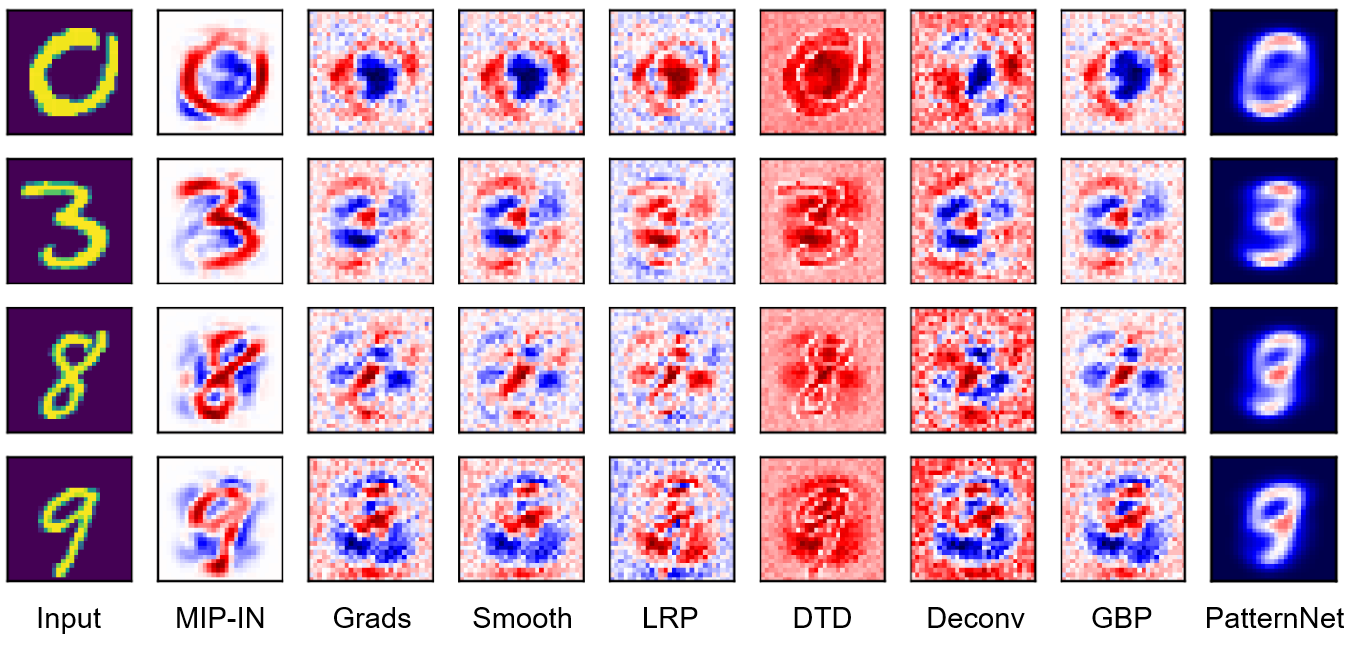}
\caption{Visualization of saliency maps comparing with seven back-propagation baselines on MNIST. MIP-IN implies that 1) the pixels on the digit have a positive effect to the final prediction, 2) while the pixels around the digit play a negative role in prediction, and 3) the regions on the background, which are scored close to $0$ and shown in white color, are not necessary for the classification.}
\vspace{-5pt}
\label{Visualization Comparison MLP MNIST}
\end{figure}

We evaluate the completeness property on three models: i) A multilayer perceptron (MLP) model trained on MNIST dataset~\cite{lecun1998mnist}; ii) A CNN model trained on MNIST; and iii) A CNN model trained on Cifar-10 dataset~\cite{krizhevsky2009learning}. The detailed network architectures are listed in Appendix. We report the APCs and Positive APCs in Table \ref{Average Percentage Change}, and also visualize the inverted source signals of several examples with tagging their classification logits in Figure \ref{Completeness visualization}. Table \ref{Average Percentage Change} shows that the APCs and especially the positive APCs, on these models are very small. Besides, the visualization in Figure \ref{Completeness visualization} also validates that the logits of the source signal are very similar to the logits of the original input signal. These evidence implies that the produced source signals well satisfy the completeness property. 

\section{Experiments}
In this section, we conduct experiments to evaluate the effectiveness of the produced interpretations by the proposed MIP-IN framework from three aspects. Firstly, we visualize the interpretation results of different model architectures on MNIST and Imagenet dataset in section \ref{Visualization results}. Secondly, we test the localization performance by comparing the produced interpretations with the bounding box annotations in section \ref{Localization performance}. Finally, we investigate the sensitivity of the interpretations with respect to class labels in section \ref{Class label sensitivity}.

\begin{figure*}[t]\centering
\includegraphics[scale=0.53]{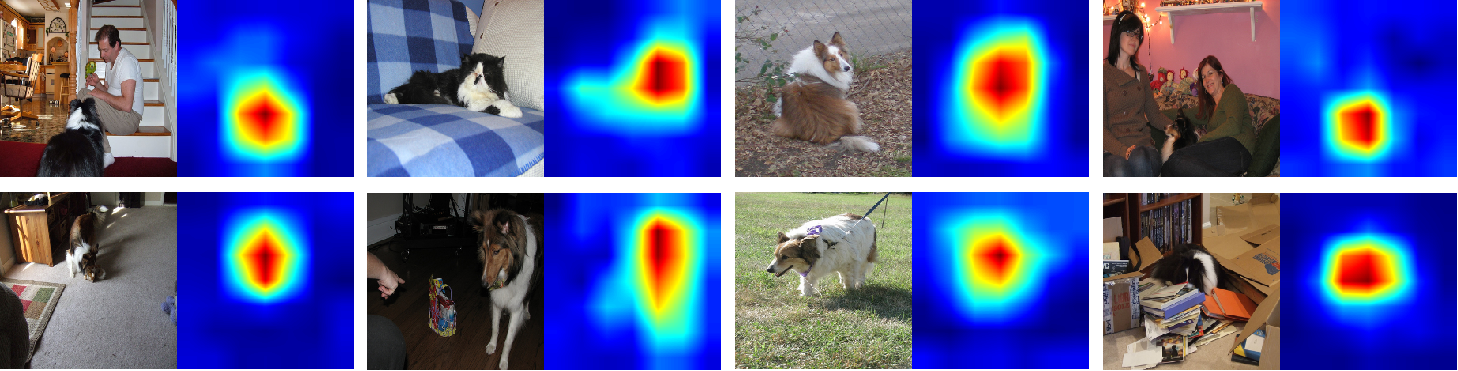}
\vspace{-5pt}
\caption{The saliency maps produced by MIP-IN  for shetland sheepdog class. Our results indicate that the VGG19 model mainly focuses on the object of interest, while ignoring the background regions. }
\label{Visualization one class vgg19}
\end{figure*}

\begin{figure*}[t]\centering
\includegraphics[scale=0.53]{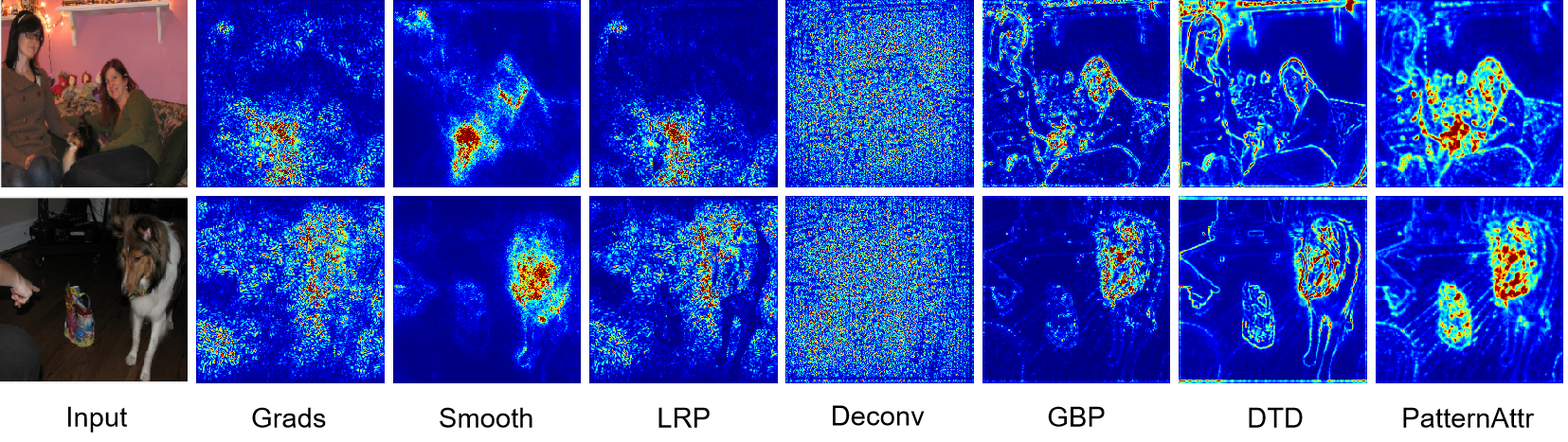}
\vspace{-5pt}
\caption{Visualization of saliency maps for seven back-propagation based baselines. They usually give some noisy scores to the background regions.}
\label{Visualization Comparison vgg19}
\end{figure*}

\subsection{Experimental Setups}
\label{Implementation}
\subsubsection{Baseline Methods}
We compare with seven popular back-propagation based attribution methods, including Grads~\cite{simonyan2014deep}, Smooth~\cite{smilkov2017smoothgrad}, LRP-$\epsilon$~\cite{bach2015pixel}, Deep Taylor Decomposition (DTD)~\cite{montavon2017explaining}, Deconvnet~\cite{zeiler2014visualizing}, GBP~\cite{springenberg2014striving}, and PatternAttribution (Patternnet)~\cite{kindermans2017learning}. Here LRP-$\alpha\beta$ is not included as it has been shown that the LRP-$0,1$ method is equivalent to DTD~\cite{montavon2017explaining}. We use the implementation of these compared methods from the innvestigate package~\cite{JMLR:v20:18-540}.

\subsubsection{Implementation Details}
We evaluate the quality of the proposed interpretations on MNIST~\cite{lecun1998mnist} and Imagenet dataset~\cite{russakovsky2015imagenet}. The training epoch for convolutional layers is set as $E = 20$. In addition, the L2 norm regularization coefficient in dense layer $\lambda = 0.001$.

On MNIST dataset, we train a three-layer MLP model on 50k training images. Then we exact the input, final prediction, and intermediate feature maps of 50k images for the model. The network parameters of MIP-IN are learned for each class recursively. The learning could be done in few minutes. On Imagenet dataset, we interpret the pre-trained VGG19~\cite{simonyan2014very} and Resnet50~\cite{he2016deep} models. The network parameters of MIP-IN are trained on the ILSVRC validation set with 50k images~\cite{russakovsky2015imagenet}. More samples for training will result in better interpretation results.
We mainly train the classifier modules. Specifically, the feature maps at top convolutional layers and the final predictions (logits) are fed as input and output respectively. Different from the classical propagation settings where forward relu stops at the second layer, the forward relu should also operate on the defined input layer (top convolutional layers). This training part takes less than one minute and repeats for $1000$ times. Then MIP-IN generates a $(512,7,7)$ dimensional attribution vector. We could derive a $(7,7)$ dimensional attribution vector by averaging over the channel dimension, then resize to original image size to obtain the saliency map, similar to Grad-CAM~\cite{selvaraju2017grad}. We apply a ReLU function to the attribution vectors for ImageNet examples because we are more interested in the features that have a positive influence on the class of interest.

\subsection{Visualization Results}
\label{Visualization results}
To qualitatively assess the interpretation performance, we visualize the saliency maps of evaluated samples of the MIP-IN and compared methods for a MLP model and the pre-trained VGG19 model.

\subsubsection{Sanity Check Measurement.}
Here we propose a novel sanity check measurement for interpretations: whether the attribution methods produce reasonable
interpretations for models without local connections or max-pooling operation, e.g., MLP models. It has been shown that the local connection largely attributes to the good visual quality of GBP~\cite{nie2018theoretical}. Besides, it's confirmed that the max-pooling operation is critical in helping DeconvNet produce human-interpretable visualizations instead of random noise~\cite{nie2018theoretical}.
However, the interpretation should be also human-interpretable without the local connection and max-pooling operation. To evaluate the performance without the distraction from local connection and max-pooling, we specially show the interpretations for a MLP model (without max-pooling) trained on MNIST in section 6.2.2.

\begin{figure*}[t]\centering
\includegraphics[scale=0.5]{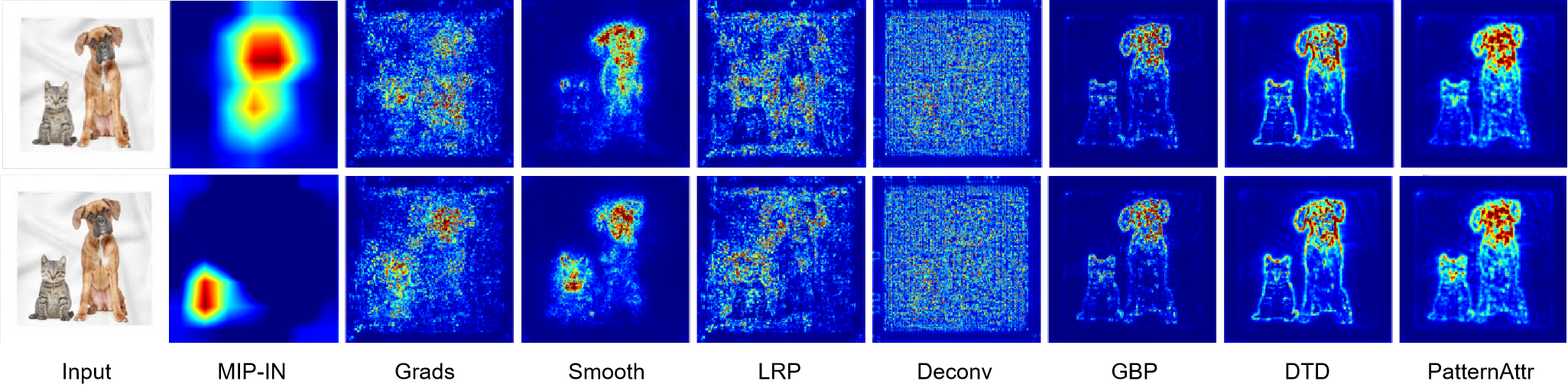}
\vspace{-6pt}
\caption{The saliency maps for dog (top row) and cat (bottom row) labels respectively. MIP-IN could generate class-discriminative interpretations. In contrast, the baseline methods produce nearly the same interpretations for both labels.}
\label{catdog}
\end{figure*}

\subsubsection{MLP Models.}
We train a three-layer MLP model on 50k training images. Figure \ref{Visualization Comparison MLP MNIST} shows the saliency maps. The proposed MIP-IN interprets the network with the best visual quality. We have the following observations. i) The attribution of compared back-propagation methods roughly localize the corresponding pixels of the digits, while the one of MIP-IN achieves an accurate localization; ii) The saliency map of compared baselines are almost noisy, which pay partial attention to the background of the image. While the visualization of MIP-IN is much sharper and cleaner than the compared baselines.
iii) The explanation of MIP-IN is consistent with human cognition. Specifically, MIP-IN implies that the pixels on the digit have a positive effect on the final prediction, while the pixels around the digit play a negative role in prediction. The regions on the background, which are scored close to $0$ and shown in white color, are not necessary for the classification. This accords with an obvious conclusion that no information of output would be embedded in these areas. in which the pixels are all constant.

\subsubsection{VGG19 Model.} We also visualize the saliency maps of eight examples in the shetland sheepdog class produced by MIP-IN in Figure \ref{Visualization one class vgg19}. Besides, the saliency maps visualizations of seven back-propagation based baselines are shown in Figure \ref{Visualization Comparison vgg19}. Figure \ref{Visualization one class vgg19} shows that the interpretations produced by MIP-IN are of good visual quality and accurately locate the object of interest. In addition, the heatmaps of GBP, DTD, and especially PatternAttrition, recover some fine-grained details of irrelevant regions, such as person. These methods suggest that the network is partially paying attention to the background. Instead, the saliency maps of MIP-IN indicate that the network is ignoring the surrounding background with almost $0$ value attributions. These observations imply that MIP-IN produces more reasonable explanations.

\begin{table}[t]
\centering
\small
\renewcommand\arraystretch{1.2}
\begin{tabular}{ C{1.9cm}|C{1.5cm}|C{1.5cm}} \hline
\textbf{Models}   & \textbf{VGG19}  & \textbf{Resnet50} \\  \hline
Grads & 0.345   & 0.368 \\ \hline
SmoothGrads & 0.518    & 0.541 \\ \hline
LRP-$\epsilon$ & 0.307   & 0.292 \\ \hline
DTD & 0.427   & 0.441 \\ \hline
Deconvnet & 0.226   & 0.263 \\  \hline
GBP &  0.398   & 0.456  \\ \hline
PatternAttr & 0.499   & - \\ \hline
GradCAM & 0.486    & 0.512 \\ \hline \hline
MIP-IN & \textbf{0.648}    & \textbf{0.662} \\ \hline
\end{tabular}
\setlength{\abovecaptionskip}{0pt}%
\setlength{\belowcaptionskip}{4pt}%
\caption{Bounding box accuracies evaluated on VGG19 and Resnet50. PatternAttribution do not support ResNet-50.}
\vspace{-3pt}
\label{Bounding box}
\end{table}

\subsection{Localization Performance}
\label{Localization performance}
To further quantitatively measure the effectiveness of the proposed interpretations, we evaluated localization performances of MIP-IN and these back-propagation based baselines. We are interested in how well these generated attributions locate the object of interest.
A common approach is to compare the saliency maps  with the bounding box annotations. Assume the bounding box contains $n$ pixels. We select top $n$ pixels according to ranked attribution scores and count the number of pixels $m$ inside the bounding box. The ratio $\alpha = \frac{m}{n}$ is used as the metric of localization accuracy~\cite{Schulz2020Restricting}.
 We follow the setting in ~\cite{Schulz2020Restricting}, which considers the scenarios where the bounding boxes cover less than $33\%$  of the input image.
We computed the localization accuracy of VGG19 and Resnet50 models on over 5k validation images, and the results are shown in Table \ref{Bounding box}. MIP-IN obtains the highest bounding box accuracy on both VGG19 and Resnet50 models. Comparing to the best baseline, MIP-IN has improved the accuracy by 0.13 and 0.121 for VGG19 and Resnet50, respectively. This substantial improvement validates that MIP-IN is consistent with human cognition.

\begin{figure}[t]\centering
\includegraphics[scale=0.38]{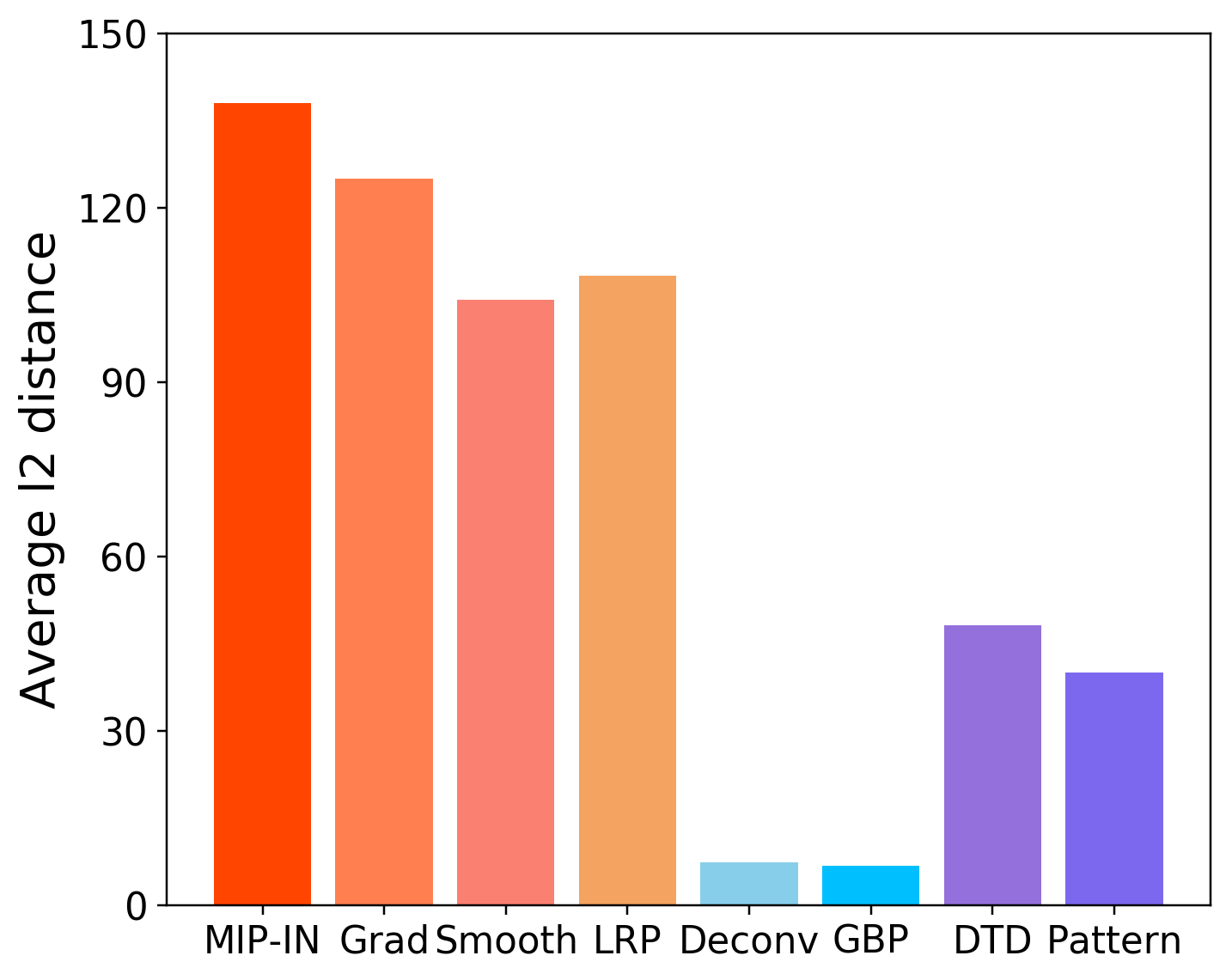}
\vspace{-7pt}
\caption{Average $l_2$ distance comparisons on VGG19 model. The large distance for two labels indicate that MIP-IN could generate class-sensitive interpretations.}
\vspace{-10pt}
\label{Average $l_2$ distance comparisons}
\end{figure}

\subsection{Class Label Sensitivity}
\label{Class label sensitivity}
A sound attribution should be sensitive to the label of the target class. Given different labels of interest, the attribution method is supposed to generate class-discriminative saliency maps. However, several works have observed that the saliency map of GBP and Decovnet keep almost the same given different class labels ~\cite{samek2016evaluating,selvaraju2017grad,mahendran2016salient,nie2018theoretical}. Hence we conduct a comprehensive investigation to the class sensitivity of the  saliency maps produced by MIP-IN and these back-propagation based baselines.

\subsubsection{Qualitative Evaluation}
To qualitatively assess the class sensitivity, we visualize the saliency maps from MIP-IN and seven baselines for the ‘bull mastiff’ (dog) class (top) and ‘tiger cat’ class (bottom) in Figure \ref{catdog}. GBP, DTD, and PatternAttribution highlight fine-grained details in the image, but generate very similar visualizations for the two classes. The observation demonstrates that these three back-propagation baselines are class-insensitive. Smooth gradients shows better sensitivity, but it highlights both dog and cat regions when interpreting the ‘cat’ class. In contrast, MIP-IN is highly class-sensitive and produce accurate interpretations. Specifically,  the ‘dog’ explanation exclusively highlights the ‘dog’ regions but not ‘cat’
regions (top row), and vice versa  (bottom row).

\subsubsection{Quantitative Evaluation}
To quantitatively describe how the back-propagation based visualizations change w.r.t. different class labels, we compute the average $l_2$ distance on
Imagenet dataset. We calculate the $l_2$ distance of two saliency maps given the class logits of two different class labels and then average these $l_2$ distances.

We evaluated the VGG19 model on $1k$ images of Imagenet. We randomly select two class labels, i.e., $981$ and $589$ class, and show the comparisons of average $l_2$ distance statistics in Figure \ref{Average $l_2$ distance comparisons}. It can be observed that the average $l_2$ distances of MIP-IN, Grad, Smooth, and LRP are much larger than Deconvnet and GBP, which demonstrate that these four attribution methods are class-sensitive while Deconvnet and GBP are not. Although DTD has a relatively larger distance than GBP and Deconvnet, we conclude that DTD is actually class-insensitive after investigating the specific examples. There are two situations for the saliency maps of DTD: i) For some examples, the two saliency maps of two classes (with both positive or negative logits) keep almost the same with zero $l_2$ distance, ii) For the other examples, the saliency map values of one class (with negative logits) are all zero values, while the ones of another class (with positive logits) are not, which results in a large $l_2$ distance. Besides, PatternAttribution is much less sensitive than MIP-IN. We also evaluated the average $l_2$ distance on a three-layer CNN model trained on MNIST, and the results are shown in Appendix.

\section{Conclusions and Future Work}
In this work, we formulate the back-propagation attribution problem as learning a source signal in the input space, which maximally preserves the mutual information between input and output. To solve this problem, we propose mutual information preserving inverse network (MIP-IN) to generate the desired source signal via globally and locally inverting the output. Experimental results validate that the interpretations generated by MIP-IN are reasonable, consistent with human cognition, and highly sensitive to the class label. In the future work, we will explore how to apply MIP-IN to generate attributions at intermediate layers and adapt MIP-IN to generate class-level attributions.



\bibliographystyle{ACM-Reference-Format}
\bibliography{main_ref}

\clearpage
\appendix
\section{Network Architectures}
In this subsection, we listed the network architectures used in the paper in Table \ref{MLP-M}, \ref{CNN-M}, and \ref{CNN-C}, respectively. The dropout technique is adopted during the training process.

\begin{table}[!h]
\centering
\renewcommand\arraystretch{1.5}
\begin{tabular}{ C{1.8cm} C{1.8cm} C{2.25cm}} \hline
\textbf{Name}   & \textbf{Activation}  & \textbf{Output size} \\  \hline
Initial &   & (28,28) \\ \hline
Dense1 &  relu   & (512,\,) \\
Dense2 & relu  & (512,\,) \\
Dense3 & softmax  & (10, \,)  \\  \hline
\end{tabular} \\
\setlength{\abovecaptionskip}{0pt}%
\setlength{\belowcaptionskip}{4pt}%
\caption{Trained MLP architecture on MNIST dataset.}
\label{MLP-M}
\end{table}

\begin{table}[htpb]
\centering
\renewcommand\arraystretch{1.5}
\begin{tabular}{ C{1.25cm} C{1.25cm} C{2cm} C{2cm}} \hline
\textbf{Name}  & \textbf{Activation}  & \textbf{kernel size}  & \textbf{Output size} \\  \hline
Initial &  &  & (28,28,1) \\ \hline
Conv1 &  relu  & (16,5,5) & (24,24,16) \\
Conv2 & relu & (64,3,3)  & (22,22,64) \\
Max1 &  & (2,2) & (11, 11, 64)  \\  \hline
Flatten & &  & (7744, \,) \\ \hline
Dense1 & relu  &  & (512, \,)  \\
Dense2 & softmax  &  & (10, \,)  \\  \hline
\end{tabular} \\
\setlength{\abovecaptionskip}{0pt}%
\setlength{\belowcaptionskip}{4pt}%
\caption{Trained CNN architecture on MNIST dataset.}
\label{CNN-M}
\end{table}

\begin{table}[htpb]
\centering
\renewcommand\arraystretch{1.5}
\begin{tabular}{ C{1.25cm} C{1.25cm} C{2cm} C{2cm}} \hline
\textbf{Name}  & \textbf{Activation}  & \textbf{kernel size}  & \textbf{Output size} \\  \hline
Initial &  &  & (32,32,3) \\ \hline
Conv1 &  relu  & (32,3,3) & (30,30,32) \\
Conv2 & relu & (64,3,3)  & (28,28,64) \\
Max1 &  & (2,2) & (14,14,64)  \\
Conv3 & relu & (64,3,3)  & (12,12,64) \\
Max2 &  & (2,2) & (6, 6, 64)  \\ \hline
Flatten & &  & (2304, \,) \\ \hline
Dense1 & relu  &  & (512, \,)  \\
Dense2 & softmax  &  & (10, \,)  \\  \hline
\end{tabular} \\
\setlength{\abovecaptionskip}{0pt}%
\setlength{\belowcaptionskip}{4pt}%
\caption{Trained CNN architecture on Cifar-10 dataset.}
\label{CNN-C}
\end{table}

\section{Average $l2$ distance on MNIST}
We also evaluated a three-layer CNN model on $5k$ images of MNIST, to see whether the number of layers would influence the conclusion. We obtain similar observations on the three-layer models, i.e., MIP-IN, Grad, Smooth, LRP are class-sensitive to the class labels with average $l_2$ distances $5.53, 7.18, 6.45, 6.88$, while GBP, DTD and PatterAttribution are not, with average $l_2$ distances $2.01, 3.62,0.91$ (DTD has the same situations to VGG19 model). Deconvnet shows a good sensitivity with $6.42$ average $l_2$ distances, which may because the trained CNN model only has one max-pooling layer.

\section{More Visualizations}
\begin{figure}[t]\centering
\includegraphics[scale=0.45]{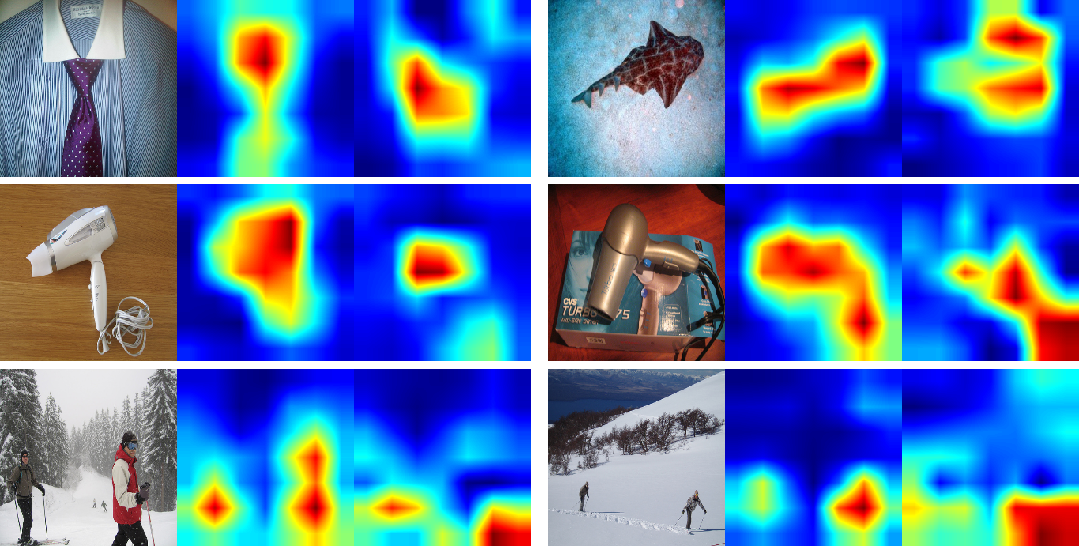}
\caption{Comparing MIP-IN with Grad-CAM.}
\label{Compared with cam}
\end{figure}

\begin{figure*}[t]\centering
\includegraphics[scale=0.6]{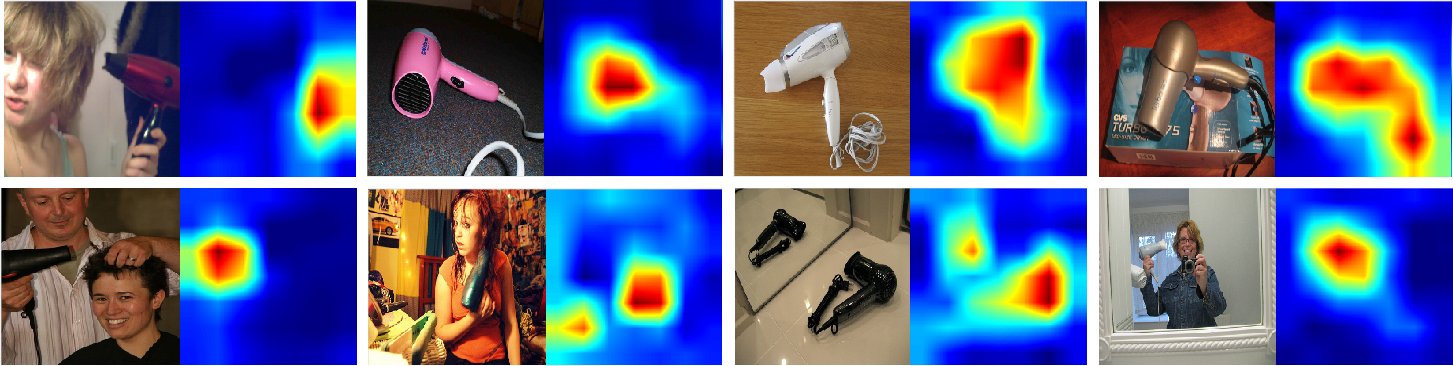}
\caption{Heatmaps of hair dryer class generated by the proposed MIP-IN method.}
\label{hairdryer}
\end{figure*}

\begin{figure*}[t]\centering
\includegraphics[scale=0.6]{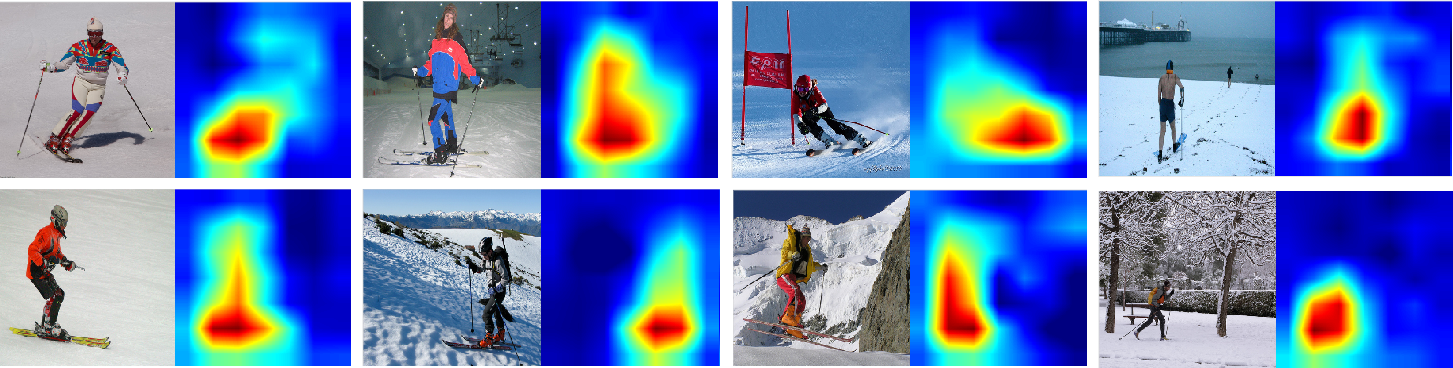}
\caption{Heatmaps of ski class generated by the proposed MIP-IN method.}
\label{ski}
\end{figure*}

\begin{figure*}[t]\centering
\includegraphics[scale=0.55]{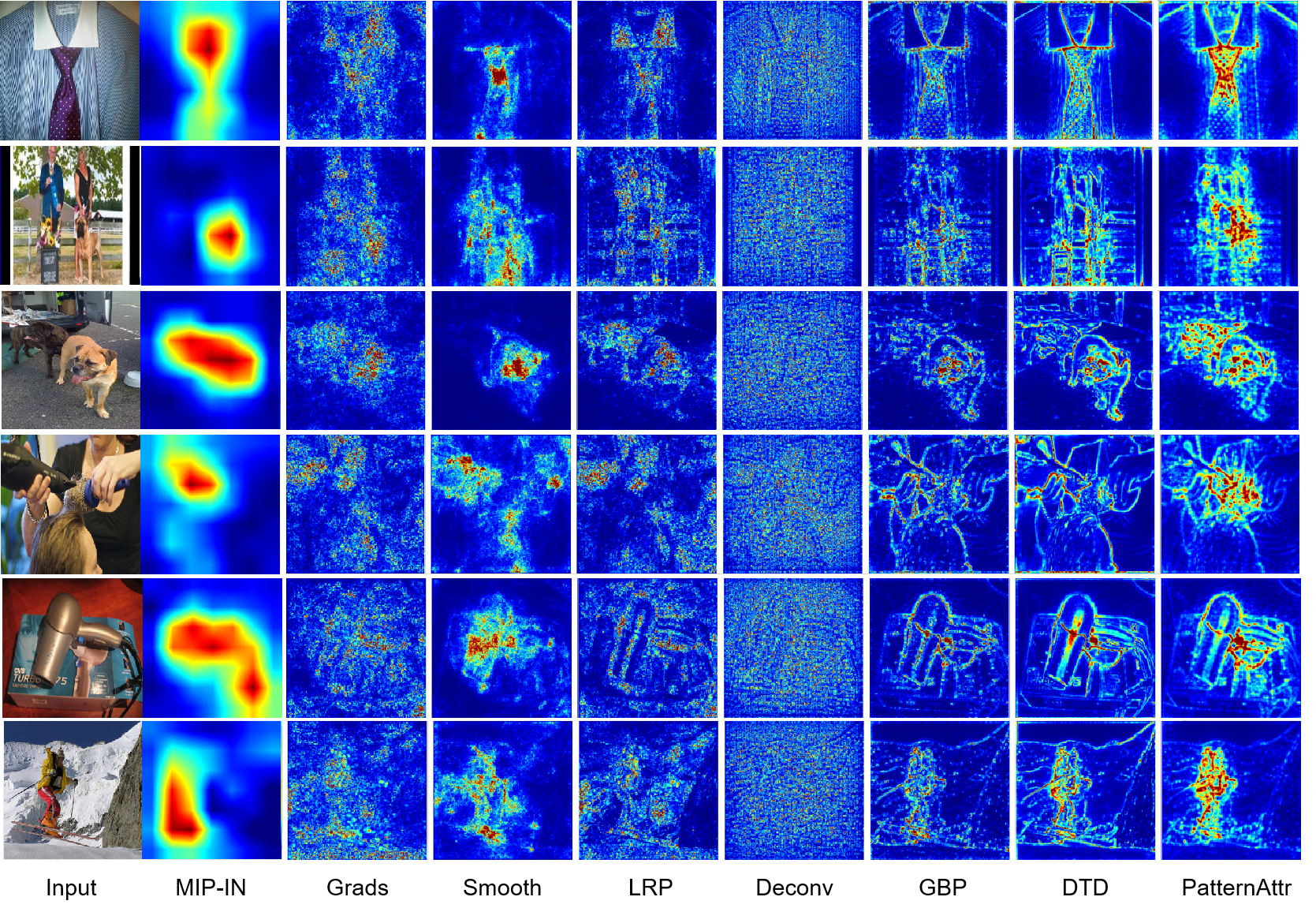}
\caption{More visualization comparison between MIP-IN and the seven baseline methods.}
\label{more-baseline-comparison}
\end{figure*}

In this section, we provide more qualitative visualizations for examples from the ImageNet dataset.
On Figure~\ref{hairdryer} and Figure~\ref{ski}, we illustrate the saliency maps for the hair dryer and ski class respectively. It indicates that MIP-IN could accurately locate the object of interest. An interesting observation is that the model would also have some attention for the human object when providing explanations for the ski class. This is perhaps humans and the skiing equipment typically co-occur in the training set, and the model would also capture this correlation and exploit it for prediction. In Figure~\ref{Compared with cam}, we also compare MIP-IN with Grad-CAM. The results indicate that MIP-IN could generate more accurate localization. In addition, we a provide comparison between MIP-IN and the seven back-propagation based attribution methods in Figure~\ref{more-baseline-comparison}.

\end{document}